\documentclass[conference]{IEEEtran}
\IEEEoverridecommandlockouts
\usepackage{cite}
\usepackage{amsmath,amssymb,amsfonts}
\usepackage{algorithmic}
\usepackage{graphicx}
\usepackage{textcomp}
\usepackage{xcolor}
\def\BibTeX{{\rm B\kern-.05em{\sc i\kern-.025em b}\kern-.08em
    T\kern-.1667em\lower.7ex\hbox{E}\kern-.125emX}}
\begin{document}
{
\title{ChartEye: A Deep Learning Framework for
Chart Information Extraction\\
}

\author{\IEEEauthorblockN{Osama Mustafa}
\IEEEauthorblockA{
\textit{Center of Excellence in AI} \\
\textit{Bahria University}\\
Islamabad, Pakistan \\
muhammadosama939@gmail.com}
\and
\IEEEauthorblockN{Muhammad Khizer Ali}
\IEEEauthorblockA{
\textit{Center of Excellence in AI}  \\
\textit{Bahria University}\\
Islamabad, Pakistan \\
mkhizer.buic@bahria.edu.pk}
\and
\IEEEauthorblockN{Momina Moetesum}
\IEEEauthorblockA{
\textit{School of Electrical Engineering and Computer Science} \\
\textit{National University of Sciences and Technology (NUST)}\\
Islamabad, Pakistan \\
momina.moetesum@seecs.edu.pk}

\and
\IEEEauthorblockN{Imran Siddiqi}
\IEEEauthorblockA{
%\textit{dept. name of organization (of Aff.)} \\
\textit{Xynoptik Pty Limited}\\
Melborne, VIC, Australia \\
imran.siddiqi@xynoptik.com.au}
% \and
% \and

}
\maketitle

\begin{abstract}
The widespread use of charts and infographics as a means
of data visualization in various domains has inspired recent research in
automated chart understanding. However, information extraction from
chart images is a complex multi-tasked process due to style variations
and, as a consequence, it is challenging to design an end-to-end system.
In this study, we propose a deep learning-based framework that provides
a solution for key steps in the chart information extraction pipeline.
The proposed framework utilizes hierarchal vision transformers for the
tasks of chart-type and text-role classification, while YOLOv7 for text
detection. The detected text is then enhanced using Super Resolution
Generative Adversarial Networks to improve the recognition output of
the OCR. Experimental results on a benchmark dataset show that our
proposed framework achieves excellent performance at every stage with
F1-scores of 0.97 for chart-type classification, 0.91 for text-role classification, and a mean Average Precision of 0.95 for text detection.
\end{abstract}


\section{Introduction}
Data visualization in the form of infographics such as charts, graphs, and plots has been widely adopted for summarization and analytical purposes in various domains. Charts also play a vital role as an integrated component of interactive visualization dashboards due to effective and efficient interpretation of complex data patterns. With the increased use of charts and plots, the need for information extraction from these has also gained popularity. Automatic information extraction from chart images is an emerging area of research that seeks to develop computer vision and natural language processing (NLP) algorithms to identify and extract data points from visual charts such as bar graphs, line plots, and pie charts and interpret their semantics. Primarily two types of information can be extracted from charts i.e. explicit and implicit. Explicit information includes the graphical and textual components that constitute a chart, while implicit information determines their relation with each other and their semantics. Explicit information extraction is the pre-requisite for implicit understanding and is the prime focus of this research as well. Due to the complexity of the problem, it is divided into several sub-modules or tasks~\cite{8978105,davila2021icpr,davila2022icpr}. The first task involves the identification of the chart type since the information layout of each type is different and may require separate processing. The second step is to detect and recognize the text and graphic components of the chart. This step is vital to determine the semantics of the chart. Another important step is the text-role classification task which identifies the labels and values of the various types of textual information present. For instance, X-axis, Y-axis, legend, titles, etc. The final step is to determine the association between the identified text roles and their values. All these tasks are characterized by challenges like variations in component position, layout, structure, text size, font, and orientation. Figure~\ref{fig:datatypes} illustrates some of the chart images to highlight the complexity of the problem. Incorrect extraction of explicit information can adversely impact the implicit inference of chart images. 

\begin{figure}[h]
\centering
\includegraphics[width=0.5\textwidth]{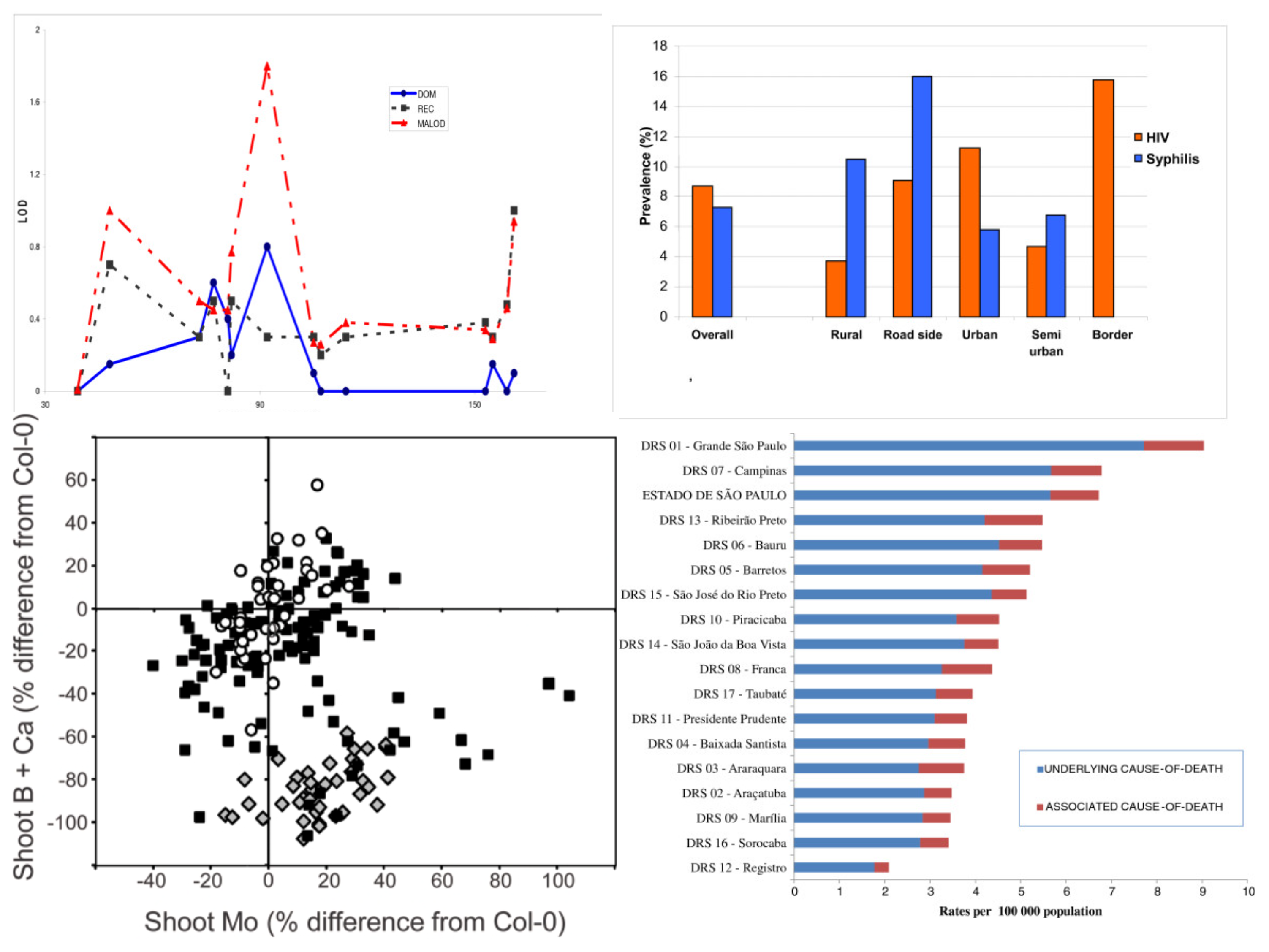}
\caption{(a) Line Plot \cite{guerrini2005genetic} (b) Vertical Bar Chart \cite{swai2006surveillance} (c) Scatter Plot \cite{baxter2009root} (d) Horizontal Bar Chart \cite{santo2012trends}}
\label{fig:datatypes}
\end{figure}

\par Due to the challenging nature of the problem, information extraction from chart images has remained an active area of research for the document analysis and recognition (DAR) community. It has also been a regular challenge at popular DAR events like ICDAR 2019~\cite{8978105}, ICPR 2020~\cite{davila2021icpr}, and ICPR 2022~\cite{davila2022icpr}. In earlier studies~\cite{karthikeyani2012machine}, traditional statistical methods (i.e. area, mean, median, and variance) were employed to classify various chart types. High level shape descriptors like Hough transform and histograms of oriented gradients (HOG) were also used for similar purposes by authors in~\cite{zhou2000bar,zhou2000hough,prasad2007classifying}. These hand-crafted features were then used to train machine learning models like Hidden Markov Model~\cite{zhou2001learning}, Multiple Instance Learning~\cite{huang2007system}, Support Vector Machines (SVM)~\cite{choudhury2016scalable,savva2011revision}, and decision trees~\cite{shukla2008recognition} to identify the input chart type. The main limitation of this approach was the lack of generalization across various types of charts. With the paradigm shift towards deep representation learning, studies like~\cite{tang2016deepchart,daggubati2022barchartanalyzer,choi2019visualizing,dai2018chart} utilized various deep learning models like deep belief networks (DBN) and convolutional neural networks (CNN) for the classification of different chart types. Most of these reported near perfect accuracies on basic chart categories like vertical and horizontal bar charts, however, their performance degraded as the number of charts increased.\\
\par For the text detection task, most of the earlier studies~\cite{al2015automatic,savva2011revision} employed traditional text detection techniques based on connected component analysis to locate textual components in chart images. Geometric (e.g. aspect ratio, width, height), structural (e.g. pixel density, edge orientation), and textural features were also used to identify the location of text in a chart image~\cite{zhou2001learning}. However, more recent studies adopted the deep object detection mechanisms for this purpose to enhance performance. For instance, authors in~\cite{poco2017reverse,poco2017extracting,dai2018chart} utilized localization methods based on CNNs for chart text detection. Two-stage deep convolutional object detectors like RCNNs have also been employed by studies like~\cite{he2017mask,liu2019data,cai2019cascade,qin2021mask}. This was a predominant trend observed in the recent competition submissions of ICPR 2020 and 2021, where most of the teams trained an RCNN variant to accomplish this task. However, most two-stage object detectors are characterized by higher time complexities as compared to one-stage architectures like YOLO. Off-the-shelf commercial tools like Tesseract OCR have also been used for text detection and recognition in chart images~\cite{smith2007overview,al2015automatic,savva2011revision}. Nonetheless, due to low resolution and small text fonts, the error rate is relatively high~\cite{dai2018chart}. Recently, attention-based mechanisms have been evaluated for chart text detection and recognition~\cite{davila2022icpr} and have shown promising results. However, their potential in this domain needs further exploration. \\ 
\par Once detected, the text needs to be categorized based on its role (e.g. legend, axis, value etc.) in the chart. This is a challenging task and requires the extraction of positional information. Wenjing DaI et al.~\cite{dai2018chart} trained an SVM using positional features for this purpose. A similar technique was proposed by Rabah et al. in~\cite{al2017machine} using random forest. Xiaoyi et al.~\cite{liu2019data} utilized a relational network (RN) for role classification in bar charts. Recently, an ensemble-based methodology was described in~\cite{davila2021icpr} that combined the output of three models trained on visual, positional, and semantic features. An accuracy of 86\% was reported.  Wang et al.~\cite{wang2021visual} utilized LayoutLM~\cite{xu2020layoutlm} for this purpose, however, the performance suffered on roles like legend title, legend label, and data markers. Mask RCNN was also evaluated and reported an accuracy of 81.7\%. Luo et al.~\cite{luo2021benchmark} extracted the positional features to train random forest and LightGBM for text-role classification and reported an accuracy of 77\%. Recently~\cite{davila2022icpr}, an RCNN with Swin transformer as backbone was also evaluated on this task and outperformed the state-of-the-art, thus validating the potential of transformers in this domain.

\par In this study, we propose a deep learning framework for explicit chart information extraction that provides a solution for each of the main steps discussed earlier i.e. chart-type classification, text detection and recognition, and text-role classification. The proposed framework introduces supplementary steps like text resolution enhancement to improve the performance of text recognition that subsequently improves the performance of the next tasks. Our proposed framework can be utilized to extract information from multiple chart-type images, thus providing a generic solution for a number of chart types. Empirical evaluations show that the proposed framework can effectively process digital images of graphical representations and output structured numerical data for further analysis. The main contributions of this study can be summarized as follows:
%Document digitization has been an area of interest among the pattern recognition community for several reasons such as document storage optimization, ease of retrieval and access \cite{sankar2006digitizing}. Digitalization of document images involves text recognition algorithms (OCR), table-to-text generation \cite{bao2018text}, data extraction from charts and so on. Many of these tools have now been matured and now available in commercial as well as open-source libraries. 

%Data extraction from charts is one of the most challenging document digitalization tools as multiple challenges are involved in this problem, due to two main reasons:  there exist many relations among objects in a chart, which is not a common consideration in general computer vision problems; and different types of charts may not be processed by the same model \cite{DBLP:journals/corr/abs-1906-11906}. The other reason is large variations in types, structure and semantics of charts 

%Typically, the legend and axes descriptions provide helpful information, such as measurement units that clarify the relationships represented by the graph. Extracting such graph components should contribute to an intelligent system that can interpret latent information in a graph. However, such components, particularly legends, are positioned in various locations, and important graph characteristics are contours and texts. Clearly, to extract graph components is difficult for traditional methods

\begin{itemize}
    \item A generic pipeline for information extraction from multiple chart-type images based on deep learning techniques. 
    \item Evaluation of the potential of latest attention-based mechanisms for chart-type classification, text recognition and text-role classification tasks.  
    \item Integration of text resolution enhancement step for improved text recognition.
\end{itemize}

\par The rest of the paper is organized as follows. Section~\ref{sec:metho} describes the proposed methodology in detail. Section~\ref{sec:exp} outlines the experimentation protocol. Section~\ref{sec:res} analyzes the results and finally, Section~\ref{sec:conc} concludes the paper.

\section{METHODOLOGY} \label{sec:metho}
In this section, we discuss the proposed framework in detail. Figure~\ref{fig:pipeline} illustrates the main steps in the pipeline i.e. chart-type classification, text detection, detected-text upscaling, text recognition, and text-role classification. The framework utilizes a combination of deep convolutional and vision transformer-based approaches to effectively extract information from chart images. For instance, for tasks such as text detection, a one-stage object detector is employed, while for chart-type classification and text-role classification, a hierarchical vision transformer is used. In order to enhance the text, an enhanced super-resolution generative adversarial network (ESRGAN) is utilized. 
We further explain the low-level architectural details of each sub-task in the subsequent sub-sections. 
\begin{figure}[!h]
\centering
\includegraphics[width=0.9\linewidth]{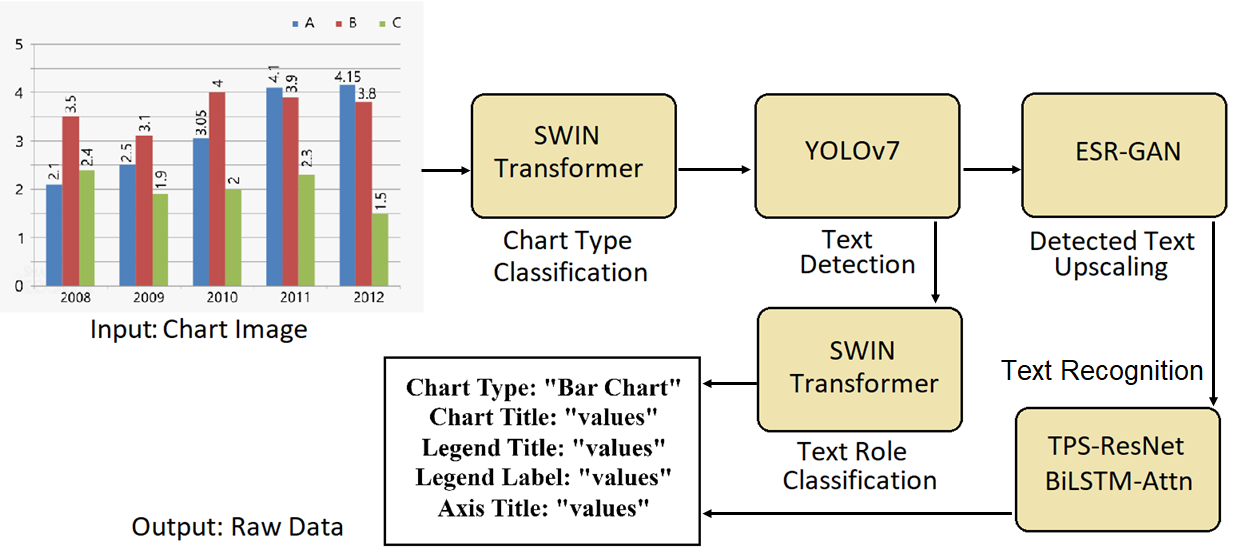}
\caption{System pipeline: A diagram illustrating how an input image is processed through different steps of the framework in a pipeline.}
\label{fig:pipeline}
\end{figure}
%Once the correct chart type is classified, in the next step text detection is performed and the output of this step is passed to the next step for upscaling and also to the last step for text-role classification, in the next step text upscaling is performed which is then passed on for text recognition, and finally, the text role is classified.  The output coming from the recognizer and text-role classifier is combined as "Role": "Value" e.g "Chart Title": "Financial Analysis Chart".
\subsection{Chart-Type Classification}
Classification of chart type is a challenging task for any classifier, as different charts have increasingly different structures and semantics. Although convolutional neural networks have shown considerable success however, there is still room for improvement in terms of performance and generalization~\cite{8978105}. In our case, we employed the state-of-the-art Swin transformer architecture~\cite{DBLP:journals/corr/abs-2103-14030}, pre-trained on ImageNet. It is a hierarchical vision transformer that uses shifted windows to generate hierarchical feature maps. Due to contextual learning, it is more robust in the classification of chart types i.e. 15 in our case as shown in Table~\ref{tab:my-tabletp}}. Figure~\ref{fig:swinseq} displays high-level architecture of Swin transformer pipeline. An input image is passed through a patch partitioning layer, the output of which enters stage 1 where a linear embedding layer and two Swin blocks are present. In stage 2, patch merging is performed and the output is passed onto the next two Swin blocks. In stage 3, patch merging with six Swin blocks is done. Finally, patch merging with two Swin blocks is repeated. 
\begin{figure}[!h]
    \centering
    \includegraphics[width=0.5\textwidth]{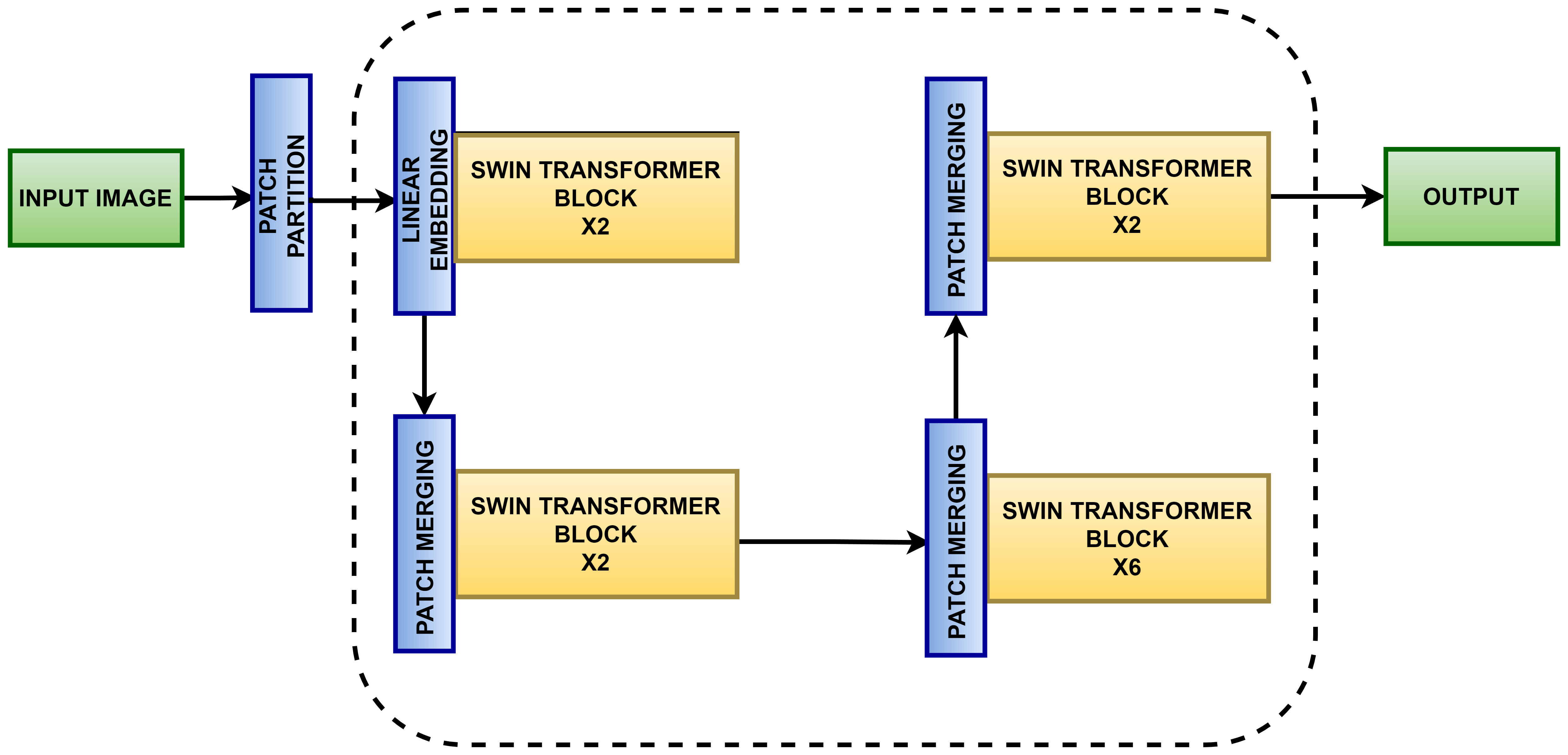}
    \caption{High-Level architecture of Swin transformer}
   
    \label{fig:swinseq}
\end{figure}
A Swin transformer block consists of a shifted window-based multi-head self-attention MSA (SW-MSA) module, followed by a multi-layered perceptron (MLP), with GELU non-linearity at in between. Layer normalization (LN) is applied before each MSA module and each MLP. A skip connection is applied after each module. A detailed architecture of two Swin transformer blocks is illustrated in the Figure~\ref{fig:SB}. With the shifted window partitioning approach, consecutive Swin transformer blocks are computed as shown in Equations~\ref{eq:1},~\ref{eq:2},~\ref{eq:3},~\ref{eq:4}.

\begin{figure}[!h]
    \centering
    \includegraphics[height=6.5cm, width=4.5cm]{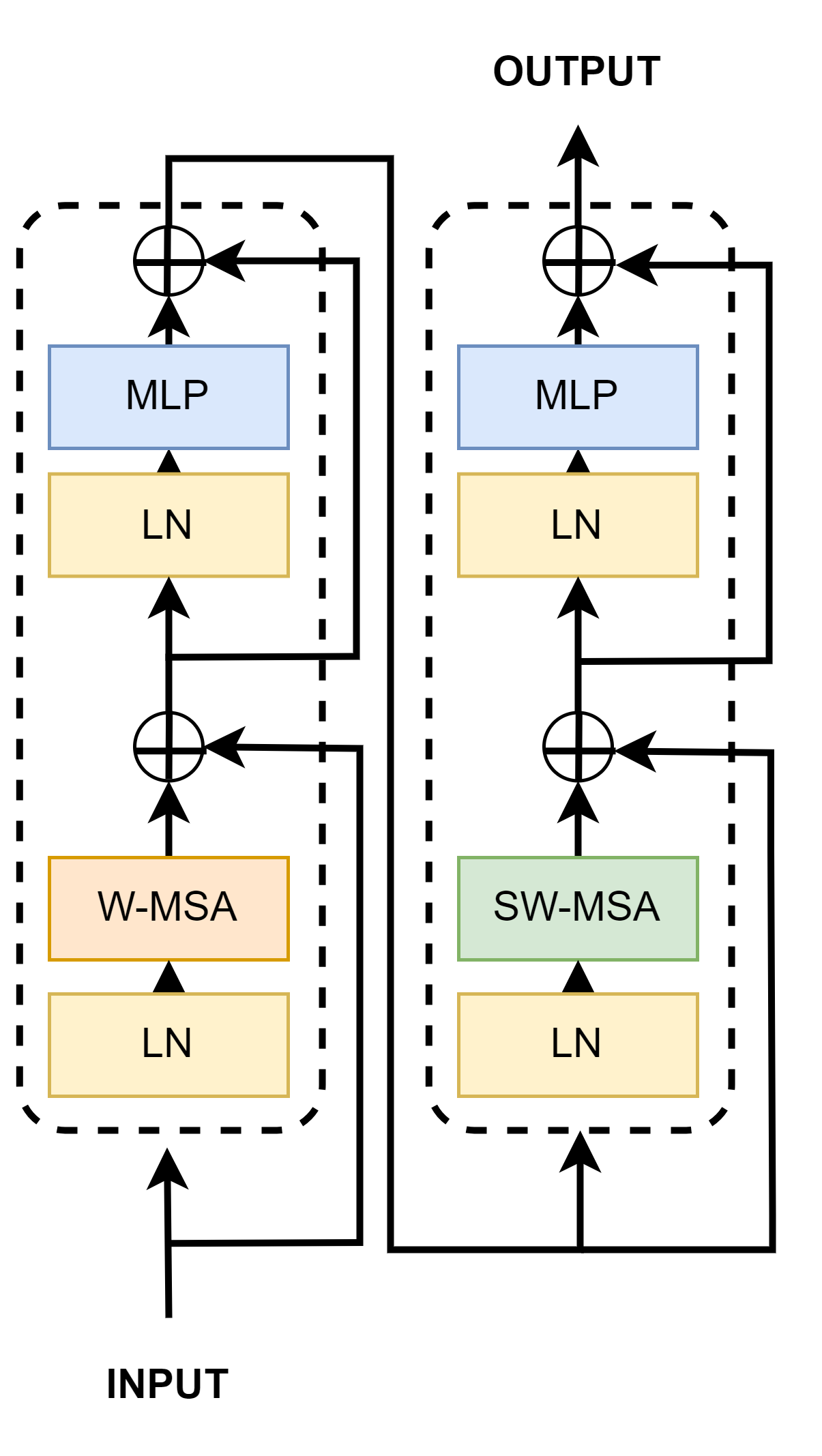}
    \caption{Architecture of two successive Swin transformer blocks. Multi-headed self-attention modules having regular and shifted windowing configuration are used i.e W-MSA and SW-MSA}
     
    \label{fig:SB}
\end{figure}

%Adapting Transformer from language to vision poses a challenge that is large variations in the scale of visual entities and the high resolution of pixels in images compared to words in text. A hierarchical transformer solves this issue and thus is more efficient as it 
%We also experimented on the same data using a ResNet-150 architecture \cite{DBLP:journals/corr/HeZRS15}, but Swin-Transformer performs better for this task. Figure~\ref{SB} illustrates the low level architecture of two consecutive swin blocks, 

\begin{equation}    
\centering
\label{eq:1}
\hat{z}^l= W-MSA(LN({z}^{l-1}))+{z}^{l-1}
\end{equation}

\begin{equation}    
\centering
{z}^l=MLP(LN(\hat{{z}}^l))+\hat{{z}}^l
\label{eq:2}
\end{equation}

\begin{equation}    
\centering
\hat{{z}}^{l+1}=SW-MSA(LN({z}^l))+{z}^l 
\label{eq:3}
\end{equation}

\begin{equation}    
\centering
z^{l+1}=MLP(LN(\hat{z}^{l+1}))+\hat{z}^{l+1} 
\label{eq:4}
\end{equation}

W-MSA and SW-MSA denote window-based multi-head self-attention using regular and shifted window partitioning configurations, respectively. $\hat{z}^l$ and ${z}^l$ denote the output features of the SW-MSA module and the MLP module for block $l$, respectively. 

%it is a multi-class classification problem with 15 chart types.  For this purpose, Our model outperforms the previous models such as CNN's and various ensemble methods used for this task as illustrated in Table 2.

\subsection{Text Detection}
Text detection is performed on the chart image, focused on logical blocks with a single role. This means that a multi-line axis-title should be detected as a single block. This step performs dual role as it not only detects text for the purpose of recognition but the detected text from this step is also passed on to the last step in pipeline for text-role classification. All types of text such as axis title, legend text, chart title, and tick values are detected. It is a challenging task where there is sometimes increasingly small-scale text present in the image. We performed transfer learning using a state-of-the-art YOLOv7~\cite{wang2022yolov7} pre-trained on MS-COCO. It is a single-stage detector that performs object detection by first separating the image into N grids of equal size. Each of these regions is used to detect and localize any objects they may contain. Although recently, transformer architectures have been utilized, in addition to some convolutional ensembles for this task~\cite{davila2021icpr}, however, YOLOv7 performs considerably in terms of precision and time-complexity.

\subsection{Detected Text Upscaling}
\begin{figure}[!h]
\centering
\includegraphics[width=1.0\linewidth]{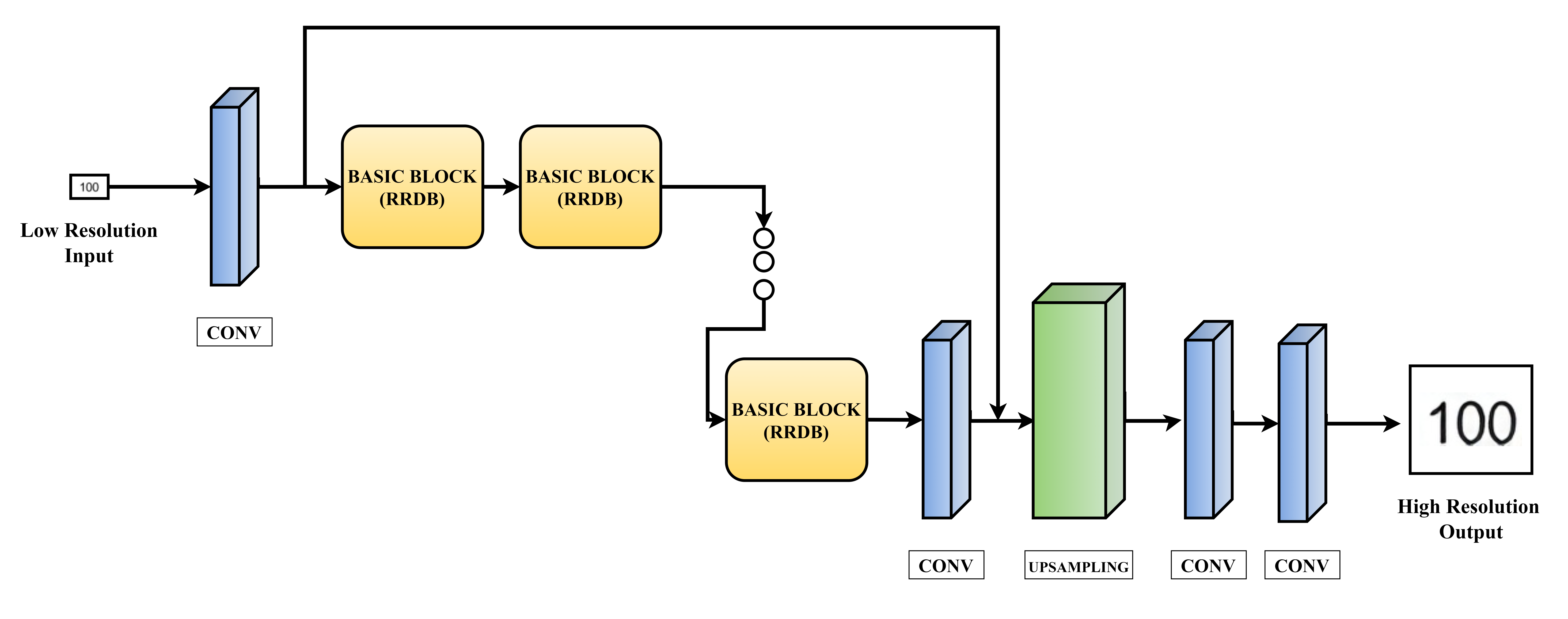}
\caption{Detected text upscaling. Pipeline utilizing enhanced super resolution GAN to enhance the resolution of low resolution input image that is detected text (The circled dots represent that there are more number of basic blocks)
}
\label{dtxupscaling}
\end{figure}

In most chart samples, the resolution of the text is very low resulting in poor recognition even by a mature text recognizer. Thus, we propose an additional step utilizing enhanced super resolution generative adversarial network (ESRGAN)~\cite{DBLP:journals/corr/abs-1809-00219} to upscale the image without losing the pixel information as compared to conventional image upscaling techniques. Image super-resolution (SR) techniques reconstruct a higher-resolution (HR) image or sequence from the observed lower-resolution (LR) images. As illustrated in Figure~\ref{dtxupscaling}, ESRGAN is an enhanced version of super resolution GAN (SRGAN), which introduces the residual-in-residual dense block (RRDB) without batch normalization as the basic network building unit. The perceptual loss has been improved by using the features before activation, which could provide stronger supervision for brightness consistency and texture recovery. The total loss for the generator is given by Equation~\ref{eq:loss}.
\begin{equation}
\centering    
L_G=L_{\text {percep }}+\lambda L_G^{R a}+\eta L_1,
\label{eq:loss}
\end{equation}
Where $L_{\text {percep }}$ is the perceptual loss, $L_G^{R a}$ is the adversarial loss for generator part in the GAN, $L_1=\mathbb{E}_{x_i}\left\|G\left(x_i\right)-y\right\|_1$ is the content loss that evaluates the 1-norm distance between recovered image $G\left(x_i\right)$ and the ground-truth $y$, and $\lambda, \eta$ are the coefficients to balance different loss terms. This proposed step helps improve the recognition accuracy in the next step significantly.

\subsection{Text Recognition}
Text recognition is performed on the up-scaled detected text from the previous step. As the text in charts is in English, thus we have utilized a TPS-ResNet-BiLSTM-Attn~\cite{DBLP:journals/corr/abs-1904-01906} architecture to perform text recognition. The model applies a thin-plate spline (TPS) spatial transformation. ResNet backbone is used as a feature extractor and bi-directional long short term memory network (BiLSTM)~\cite{DBLP:journals/corr/HuangXY15} for sequence modeling. BiLSTM is a type of RNN architecture that processes input sequences in both forward and backward directions, handling variable-length input sequences and learning context from both past and future inputs making it good for sequence modeling tasks. Finally, an additive attention mechanism is employed for the prediction part allowing the model to focus on specific parts of the input that are most relevant for making predictions. %Lastly, an attention mechanism for the prediction part, allows the model to focus on specific parts of the input that are most relevant for making predictions. #old

\subsection{Text Role Classification}
\begin{figure}[!h]
\centering
\includegraphics[width=0.8\linewidth]{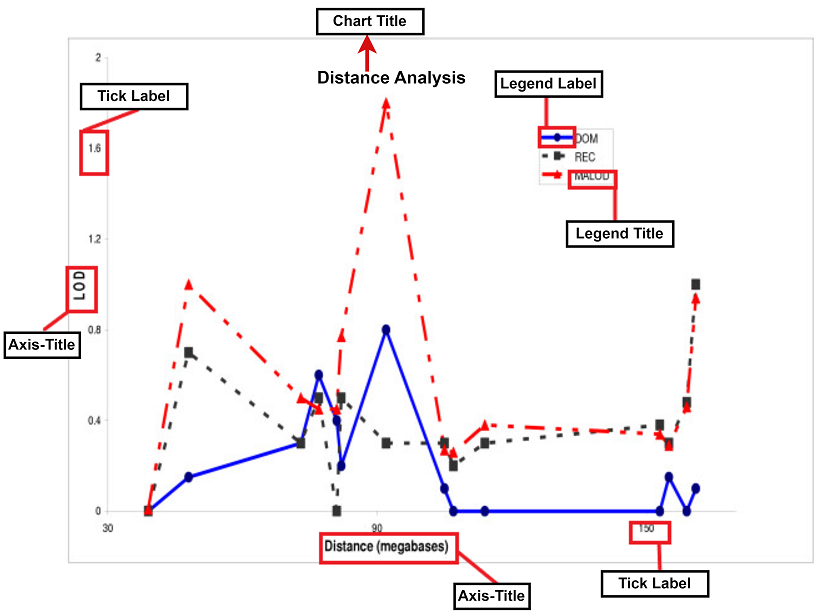}
%\caption{Text roles on a line plot} #old caption
\caption{An illustration of different text-roles on a line plot chart type.}
\label{fig:my_labelcannot}
\end{figure}

Text on a chart image has a specific role. The purpose of this task is to classify the role of text. Nine roles are being considered in this study. These include chart title, mark label, legend title, legend label, axis title, tick label, tick grouping, value label, and others. It is a challenging task due to the position, orientation, and size variations of text in a chart~\cite{8978105,davila2021icpr}. Figure~\ref{fig:my_labelcannot} shows how different texts and their roles appear on a line plot. In this study, we attempt to address these challenges by fine-tuning a Swin transformer~\cite{DBLP:journals/corr/abs-2103-14030} as a separate classifier for text-role classification instead of using YOLOv7's recognizer. Therefore, the text detected by the YOLOv7 model is fed to the Swin transformer for role classification after upscaling. This enhances the performance of the text-role classification step in different chart types. Also Swin Transformer's ability to capture multi-scale features, positional encoding, and self-attention mechanism makes it an effective technique for handling orientation issues in chart images. We also performed experimentation using DEtection TRansformer (DETR)~\cite{DBLP:journals/corr/abs-2005-12872} and YOLOv7. However, Swin trasformer outperforms both of these techniques significantly. 

\section{Experimental Protocol} \label{sec:exp}
This section details the experimental protocol employed in this study. The main objective of our experiments is to validate the methodology employed in each step. The scope of the study is not to propose an end-to-end system but instead provide a framework. Hence, each step is independently evaluated. In an end-to-end system, the performance of each subsequent step is dependent on the previous tasks and therefore it is hypothesized that performance improvement of one step will enhance the overall accuracy of the complete system. In the subsequent subsections, we provide details of the dataset employed followed by model training and testing for each task i.e. chart-type classification, text detection, text recognition, text enhancement, and text-role classification. 
\subsection{Dataset}
To evaluate the performance of each task, we employed the ICPR2022 CHARTINFO UB PMC competition dataset~\cite{davila2022icpr}. The dataset provides chart images for 15 different chart types along with annotations for each image. This version of dataset has been used in ICPR 2022 challenge. It contains real charts extracted from Open-Access publications found in the PubMed Central repository. The dataset contains only real chart images, no synthetic data is included in this version. However, the number of samples for each chart type is not balanced as shown in Table~\ref{tab:my-tabletp}.
\begin{table*}[!h]
\centering
\caption{Dataset Details for Text-Role Classification}
\label{tab:my-tableccount}
\resizebox{\linewidth}{!}{%
\begin{tabular}{lllllllllll}
\hline
Chart Type     & Images & mark-label & chart-title & legend-title & other & tick-grouping & axis-title & legend-label & tick-label & value-label \\
\hline
Horizontal Bar & 787 & 2          & 83          & 14           & 733   & 200           & 647        & 1231         & 12504      & 3725        \\
Vertical Bar   & 5454 & 57         & 118         & 34           & 1312  & 434           & 2838       & 3465         & 26441      & 3082        \\
Line Plot      & 10556 & 1263       & 143         & 69           & 2064  & 38            & 4406       & 5144         & 32462      & 667         \\
Scatter Plot   & 1350 & 589        & 57          & 52           & 1226  & 10            & 1832       & 1952         & 13281      & 55          \\
\hline 
\end{tabular}}
\end{table*}

\begin{table}[!h]
\footnotesize
\centering
\caption{DATASET DETAILS FOR CHART-TYPE CLASSIFICATION}
\label{tab:my-tabletp}
\begin{tabular}{llll}
\hline
Chart Type     & No. of Images & Chart Type            & No. of Images         \\ \hline
Horizontal Bar & 787           & Horizontal Interval   & 156                   \\
Vertical Bar   & 5454          & Manhattan             & 176                   \\
Line Plot      & 10556         & Map                   & 533                   \\
Scatter Plot   & 1350          & Pie                   & 242                   \\
Scatter-Line   & 1818          & Surface               & 155                   \\
Vertical Box   & 763           & Venn                  & 75                    \\
Area           & 172           & Vertical Interval     & 489                   \\
Heatmap        & 197           & - & - \\
\hline
\end{tabular}
\end{table}
\par Moreover, for text-role classification task, we have considered only four types of charts as the data annotation for text components in these chart types is sufficient for experimental protocol as compared to other classes. The number of samples for each text-role class in these four chart types are outlined in Table~\ref{tab:my-tableccount}. 
%\intextsep

\subsection{Model Training}
All models are trained on a NVIDIA Tesla P100 16GB GPU hardware. 80\% of the data for each task is employed for model training while 20\% is used for testing purposes.
\begin{itemize}
\item Firstly, we train the chart-type classification model. A pre-trained Swin transformer (swin-large 224~\cite{DBLP:journals/corr/abs-2103-14030}) is employed for this purpose. There are 195M trainable parameters and the train-time is 12h. The model is fine-tuned on our dataset for 50 epochs using a batch-size of 8. Adam optimizer is used with a learning rate of lr = 0.000003.  Sparse-categorical cross-entropy loss is computed during training. We also evaluated the performance of a pre-trained ResNet150 model, but the results achieved using Swin transformers outperform those obtained by ResNet150.\\ 
\item For text detection task, we applied transfer learning using a pre-trained YOLOv7 base. There are 36.9M trainable parameters and the average train-time is 16h. The model is trained for epochs: 70 (horizontal bar charts), 100 (vertical bar), 50 (line plot), and 50 (scatter plot) before convergence. A batch-size of 4 is used along with Adam optimizer at a learning rate of lr = 0.000005. Additional hyperparameters include 
 a mutation scale ranging between 0-1, momentum values of 0.3, 0.6, and 0.98, and weight-decay of 1, 0.0, and 0.001.\\
\item Output of YOLOv7 is then enhanced using a variant of enhanced super resolution generative adversarial network (ESRGAN). An upscaling of 1.5x is applied. We experimented with different upscaling values and observed that 1.5x is the optimal in most cases. Increasing this value distorted the shape of characters and affected the recognition module's performance. Figure~\ref{uchk}  shows a detected text from chart image, upscaling it to 1.5x improves the resolution, and upscaling to 3.0x deforms the shape of letter `e'.\\

\begin{figure}[!h]
    \centering
    \includegraphics[width=0.8\linewidth]{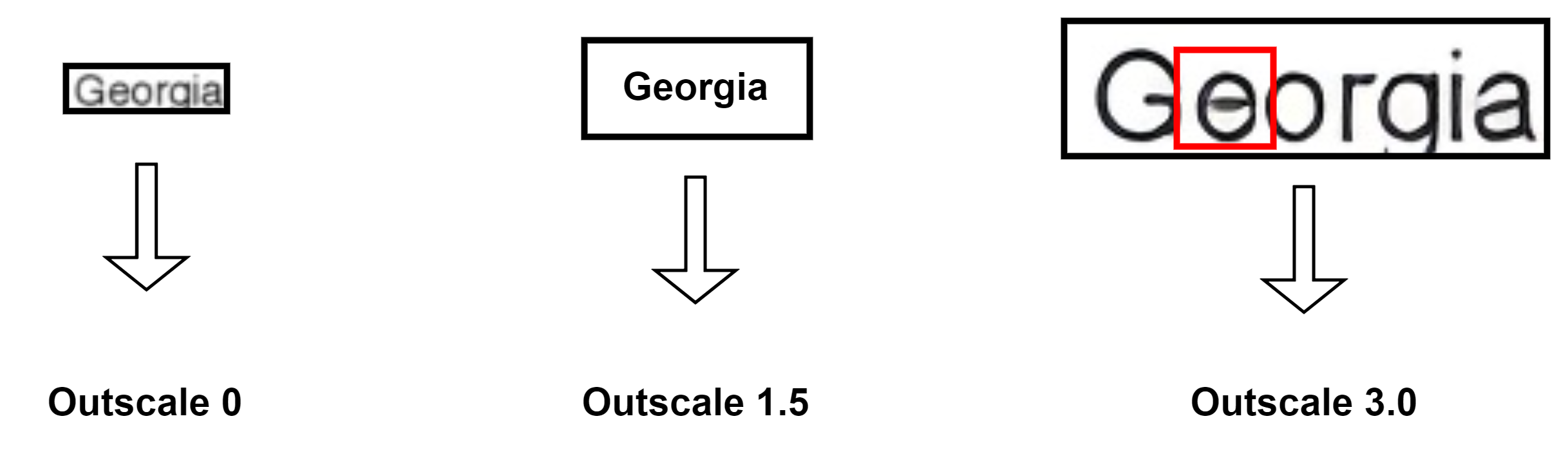}
    %\caption{Upscaling results on different outscale parameter values} #old caption
    \caption{Results of detected-text resolution enhancement using ESRGAN on different outscale parameter values. If the outscale parameter value is 1.5, the resolution of the input image will be enhanced to 1.5x its original resolution. }
    \label{uchk}
\end{figure}

\item Text recognition is performed on the output of the previous step, which is the upscaled version of detected text. For recognition purposes, we employed TPS-ResNet-BiLSTM-Attn model. The architecture utilizes a thin-plate spline transformer (TPS) to normalize the input
text image. Text components in chart images usually come in diverse shapes. For recognition of such text images, a recognizer requires to learn an invariant representation with respect to geometry. To ease this process TPS employs a smooth spline transformation between a set of fiducial points. This provides flexibility of various aspect ratios. Normalized text is then mapped into feature space by a ResNet backbone. The extracted features are reshaped into a sequence and fed to a BiLSTM for modeling. An attention-based sequence prediction mechanism is then used to predict the text.\\
% \item Finally, we train our text-role classification model. Text-role classification proved to be one of the most challenging task in the entire pipeline. Text roles being considered in our study are chart-title, legend-title, legend-label, axis-title, tick-label, tick-grouping, mark-label, and value-label. For every other text, `other' category is used. We experimented with three state-of-the-art architectures for this task. First we experimented with DEtection TRansformer (DETR)~\cite{DBLP:journals/corr/abs-2005-12872} architecture that is initially pre-trained on MS-COCO. Transfer learning is performed on our dataset. The model is trained for 100 epochs. Adam optimizer is applied with a learning rate of le-5 for the transformer, and le-6 for the backbone. We also evaluated the performance of YOLOv7~\cite{wang2022yolov7} architecture also pre-trained on MS-COCO. Again transfer learning on our data is performed as per the following train configurations i.e. epochs: 100 (horizontal-bar, vertical-bar), 50 (line-plot), 50 (scatter-plot), batch-size: 4, lr: 0.000005, and optimizer: `Adam'. Lastly, we performed transfer learning on a Swin transformer and used it as a classifier. There are 195M trainable parameters and the train-time is 14h. The model was trained for 50 epochs using batches of size 8 and a learning rate of lr = 0.000003. For all trainings, we employed cropped bounding box images from our dataset ground truth. Figure~\ref{trccurves} shows the impact of increasing the number of epochs on the classification accuracy.\\

\item We trained our text-role classification model, considering roles like chart-title, legend-title, legend-label, axis-title, tick-label, tick-grouping, mark-label, and value-label; using three state-of-the-art architectures: DETR~\cite{DBLP:journals/corr/abs-2005-12872} (100 epochs), YOLOv7~\cite{wang2022yolov7} (100 epochs for horizontal-bar and vertical-bar, 50 epochs for line-plot and scatter-plot), and Swin transformer (50 epochs). We employed batch sizes of 4 (YOLOv7) and 8 (Swin transformer) with respective learning rates. All models used cropped bounding box images from our dataset ground truth. Figure~\ref{trccurves} depicts the impact of increasing epochs on classification accuracy.\\
\begin{figure}[!h]
\begin{tabular}{cc}
  \includegraphics[width=37mm]{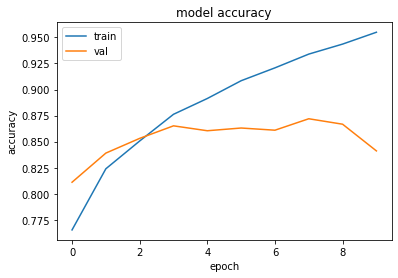} &   \includegraphics[width=37mm]{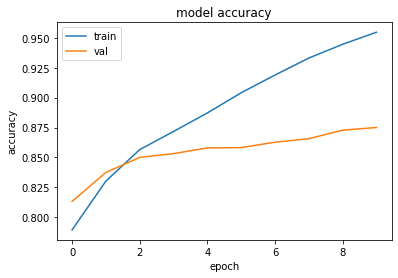} \\
(a) Horizontal Bar Chart & (b) Vertical Bar Chart \\[6pt]
 \includegraphics[width=37mm]{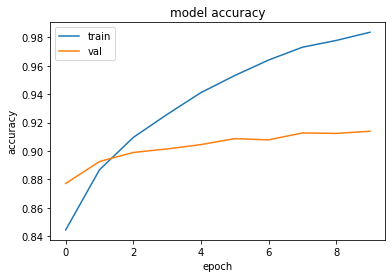} &   \includegraphics[width=37mm]{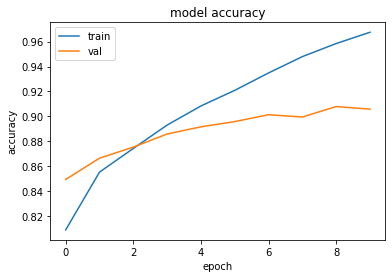} \\
(c) Line Plot & (d) Scatter Plot \\[6pt]

\end{tabular}
\caption{Text Role Classification Train-Validation Accuracy Curves}
\label{trccurves}
\end{figure}
\end{itemize}

\subsection{Evaluation Metrics}
For each task in the pipeline, we employ the following evaluation metrics. For chart type classification, we use Accuracy, Precision, Recall, and F1-score (given in Equation~\ref{eq:6},\ref{eq:7},\ref{eq:8},\ref{eq:9}). For text detection module, we use mean Average Precision (mAP). Here mAP refers to mAP50 which is calculated by computing the average precision at only the 50\% IoU threshold. For text role classification, we used F1-Score measure.\\
\begin{equation}
Accuracy = \frac{TP+TN}{TP+TN+FP+FN}
\label{eq:6}
\end{equation}
\begin{equation}
Precision = \frac{TP}{TP+FP}
\label{eq:7}
\end{equation}
\begin{equation}
Recall = \frac{TP}{TP+FN}
\label{eq:8}
\end{equation}
\begin{equation}
F1 = \frac{2*Precision*Recall}{Precision+Recall} = \frac{2*TP}{2*TP+FP+FN}
\label{eq:9}
\end{equation}\\
Where TP, TN, FP, and FN are the True Positives, True Negatives, False Positives, and False Negatives, respectively.

\section{Results and Discussion} \label{sec:res}
\begin{figure*}[!h]
\centering
\includegraphics[width=1.0\linewidth]{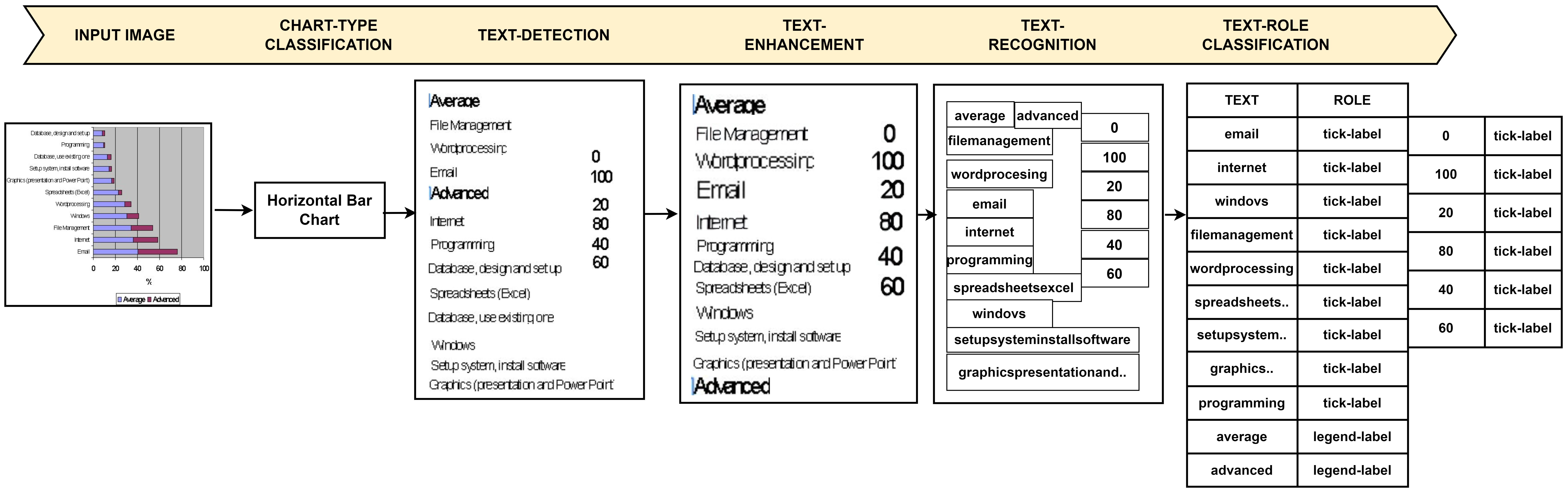}
\caption{An input chart image processed through all steps of the framework.
}
\label{overallflow}
\end{figure*}

In this section, we discuss the outcomes of our proposed methodologies for each task in the information extraction pipeline. Figure~\ref{overallflow} illustrates an input chart image that is being processed through all steps of the framework pipeline. 
\newline

%rigorous experimentation has been conducted in each task of this proposed framework. The experimentation results validate that it outperforms SOTA in each task. The training graphs and tables presented in this section helps to analyze the performance of this proposed framework. Furthermore we discuss the results of each task respectively.
\subsection{Chart Type Classification Results}
In chart type classification, our model achieves an F1-score of 0.97 and as it can be seen in Table~\ref{tab:my-tablectc}, the achieved F1-score by our model is higher than the previous techniques used for this task such as ResNet, VGG, and Swin ensemble. Although the test set used for evaluation of these SOTA techniques was different from what we have used so direct comparison is not an option, but this reference acts as an indicator that our model performs well for this task. The model is able to correctly classify 15 chart types with an overall F1-score of 0.97 which validates that it is able to handle the style variations and other challenges mentioned earlier. The results also validate that the used hierarchical vision transformer in this task is successfully able to achieve deep position-level learning to obtain significant results. \\
\begin{table}[!h]
\centering
\caption{Performance Comparison on Chart-Type Classification}
\label{tab:my-tablectc}
\resizebox{0.45\textwidth}{!}{%
\begin{tabular}{lll}
\hline
\textbf{Reference} & \textbf{Architecture}                & \textbf{F1-Score} \\
\hline
\textbf{Proposed}  & \textbf{Swin Transformer (large224)} & \textbf{0.970}     \\
ICDAR2019 \cite{8978105}         & ResNet-101                           & 0.882             \\
ICDAR2019 \cite{8978105}         & VGGNet                               & 0.775             \\
ICPR2020 \cite{davila2021icpr}          & DenseNet121+ResNet152                & 0.928             \\
ICPR2020 \cite{davila2021icpr}          & ResNet50                             & 0.904             \\
ICPR2022 \cite{davila2022icpr}          & Swin224+Swin384                      & 0.910             \\
\hline
\end{tabular}}
\end{table}
\subsection{Text Detection and Recognition Results}
For text detection, our fine-tuned YOLOv7 model is able to successfully detect text on all four chart types under consideration. The results are outlined in Table~\ref{tab:my-tablectc14a}. The results validate that YOLOv7 in this task performs well along with other SOTA techniques utilized in recent literature. 
\begin{table}[!h]
\centering
\caption{Text Detection Results}
\label{tab:my-tablectc14a}
\resizebox{0.5\linewidth}{!}{%
\begin{tabular}{ll}
\hline
\textbf{Chart Type} & \textbf{mAP} \\
\hline
\multicolumn{1}{l|}{Horizontal Bar Chart} &  0.953                   \\
\multicolumn{1}{l|}{Vertical Bar Chart}   & 0.972                   \\
\multicolumn{1}{l|}{Line Plot}            & 0.960                   \\
\multicolumn{1}{l|}{Scatter Plot}         & 0.925  \\
\hline
\end{tabular}}
\end{table}\\
\par
\begin{figure}[!h]
    \centering
    \includegraphics[width=1.0\linewidth]{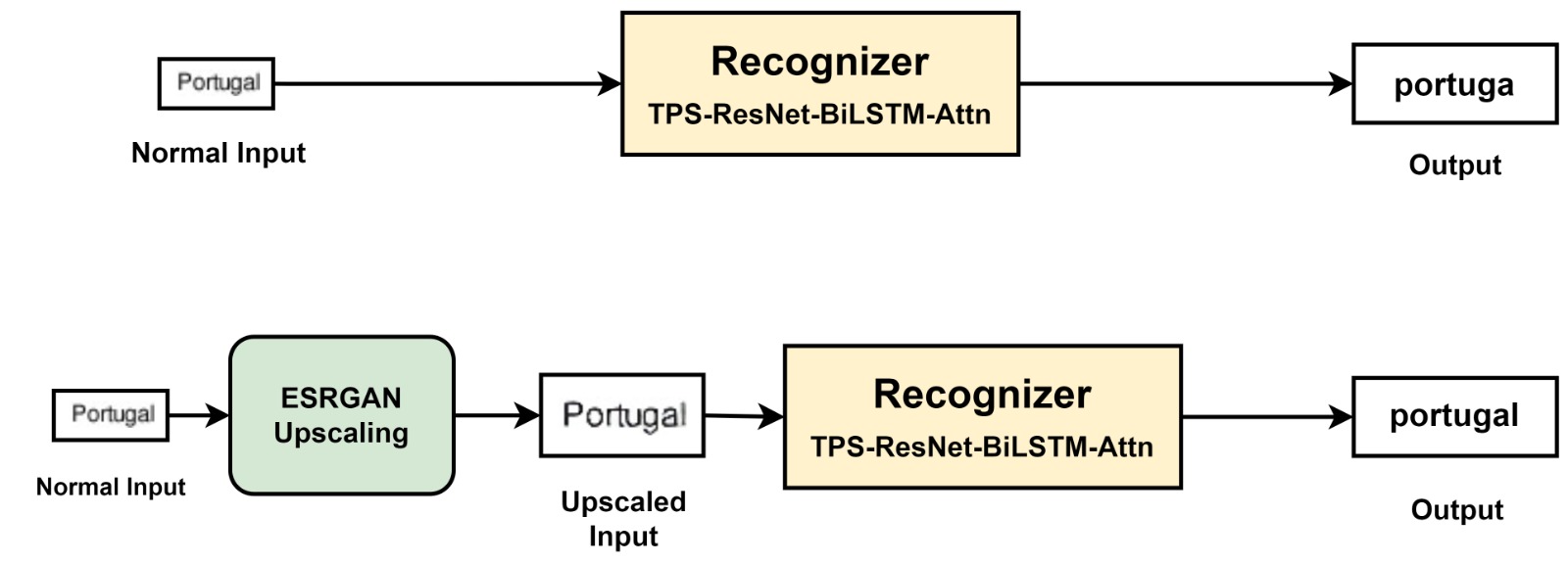}
    %\caption{Text recognition results before and after ESRGAN upscaling} #old caption
    \caption{Text-recognition results before \& after resolution enhancement-ESRGAN.}
    \label{fig:recogqualify}
\end{figure}

 As mentioned earlier, due to the complex layout and variations of font styles and sizes, conventional OCRs do not perform well on chart text recognition. For this purpose, we evaluated the performance of state-of-the-art TPS-ResNet-BiLSTM-Attn model that has shown to outperform popular scene text recognition algorithms. As expected, the model is able to perform significantly better than conventional OCRs on chart text as well. However, despite its success on most text types, it did not perform well on very small-sized text as shown in Figure~\ref{fig:recogqualify}. To cater for this issue, we introduced an additional step of enhancing resolution of detected text before recognition. Figure~\ref{fig:recogqualify} shows the improvement in recognition module after enhancement.

\subsection{Text Role Classification Results}
Text style variations and relations between multiple text objects make text-role classification a challenging task for any classifier. Furthermore, placement of a specific type of text is also highly dependent on the chart type. Thus making it difficult to device a generic tool for different types of charts. In this study, we are considering nine role classes across four different chart types. As illustrated in Table~\ref{tab:my-tablectc14b}, our trained hierarchical vision transformer using shifted windows is able to achieve considerable success in this task for all chart types. The model achieves an overall F1-Score of 0.91, 0.90, 0.86, and 0.84 for all nine text-role classes in line plot, scatter plot, vertical bar, and horizontal bar charts, respectively.  \\
\begin{table}[!h]
\centering
\caption{Text-role Classification Results}
\label{tab:my-tablectc14b}
\resizebox{0.45\textwidth}{!}{%
\begin{tabular}{lllll}
\hline
\textbf{Chart Type}                       & \multicolumn{4}{c}{\textbf{Text Role Classification}}                                             \\
\hline
\multicolumn{1}{l|}{\textbf{}}            & \textbf{Accuracy} & \textbf{F1 Score} & \textbf{Precision} & \multicolumn{1}{l}{\textbf{Recall}}  \\
\cline{2-5}
\multicolumn{1}{l|}{Horizontal Bar Chart} & 0.870             & 0.842            & 0.848             & 0.841   \\
\multicolumn{1}{l|}{Vertical Bar Chart}   & 0.875             & 0.866            & 0.866            & 0.875   \\
\multicolumn{1}{l|}{Line Plot}            & 0.914             & 0.909            & 0.911            & 0.913\\
\multicolumn{1}{l|}{Scatter Plot}         & 0.905            & 0.902            & 0.903             & 0.905 \\
\hline
\end{tabular}}
\end{table}\\
\par To understand the complexity of problem and the significance of our methodology, we compare our results with two state-of-the-art architectures YOLOv7 and DETR, where mAP is 0.40 and 0.30 from these techniques respectively. Thus, pre-trained Swin transformer fine-tuned on our data outperforms both YOLOv7 and DETR with significant margin. 
% \begin{table}[!h]
% \centering
% \caption{Performance comparison on Text-role Classification}
% \label{tab:my-tabletrccomparison}
% \resizebox{0.45\linewidth}{!}{%
% \begin{tabular}{ll}

% \hline
% Technique                    & mAP  \\
% \hline
% DEtection TRansformer (DETR) & 0.30 \\
% YOLOv7                       & 0.40 \\
% \textbf{Proposed: SWIN Transformer}    & \textbf{0.90} \\
% \hline
% \end{tabular}}
% \end{table}\\

\par
\begin{figure}
\centering
\includegraphics[width=0.5\textwidth]{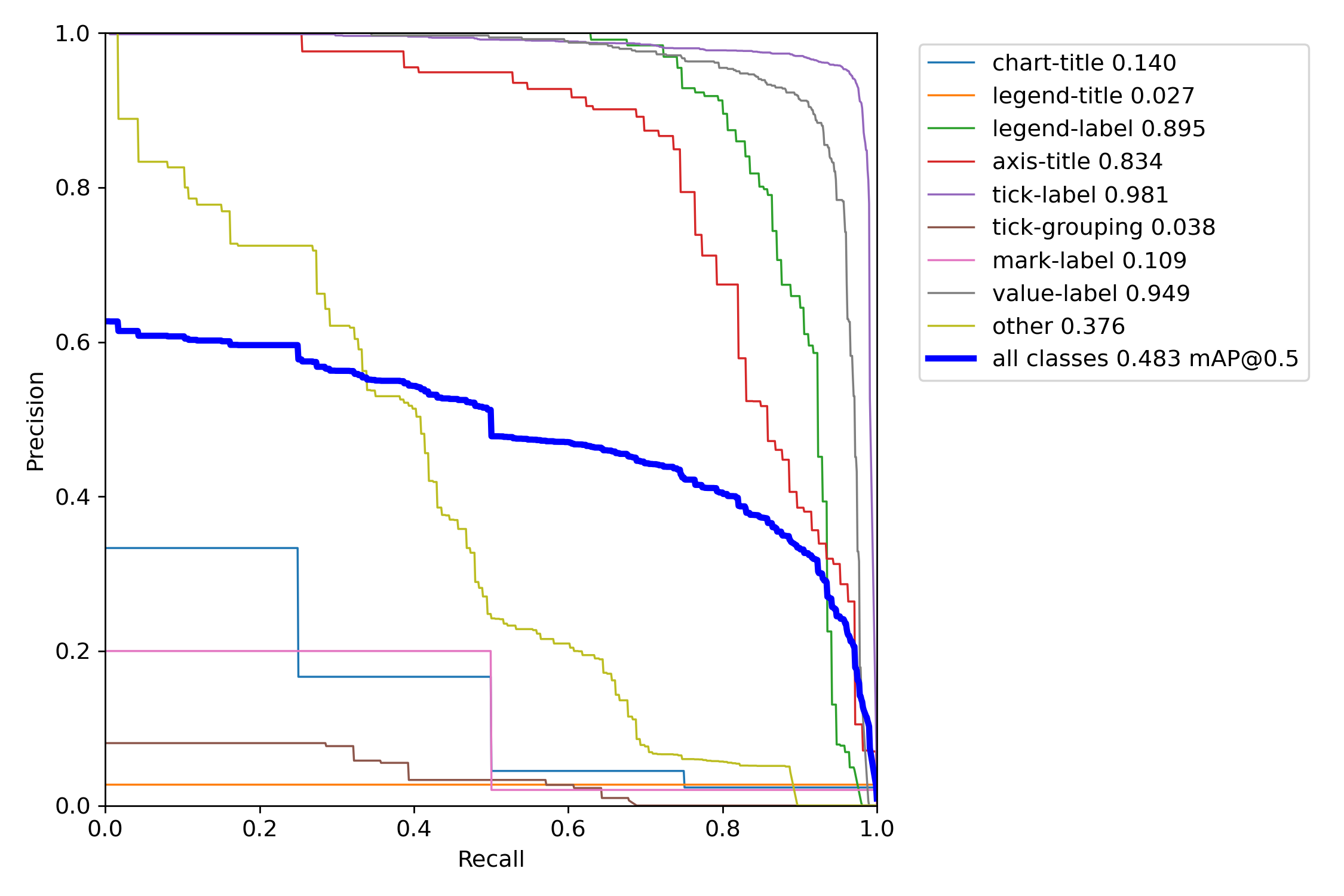}
\caption{YOLOv7-PR curve for text-role classification }
\label{prcurve}
\end{figure}

 Although YOLOv7 performed well in text detection but its performance in text-role classification is not satisfactory. Figure~\ref{prcurve} shows the results of YOLOv7 for all text-role classes. Except for legend-label, axis-title, tick-label, and value-label, the model performs below satisfactory for the rest of the classes. On the other hand, Swin transformer improves overall accuracy of the text-role classification significantly.  \\
\\

% \begin{table}[!h]
% \centering
% \caption{Text Role Classification Results}
% \label{tab:my-tabletrcres}
% \resizebox{0.5\linewidth}{!}{%
% \begin{tabular}{lllll}
% \hline
% \hline
% Chart Type           & Accuracy & \textbf{F1-Score} & Precision & Recall \\
% \hline
% Horizontal Bar Chart & 0.870    & \textbf{0.8422}   & 0.8481    & 0.8411 \\
% Vertical Bar Chart   & 0.875    & \textbf{0.8669}   & 0.8666    & 0.8750 \\
% Line Plot            & 0.914    & \textbf{0.9097}   & 0.91111   & 0.9138 \\
% Scatter Plot         & 0.9058   & \textbf{0.9026}   & 0.9038    & 0.9057 \\
% \hline

% \end{tabular}}

% \end{table}

% \begin{table}[!h]
% \centering
% \caption{Text Detection Results}
% \label{tab:my-tabletd4}
% \resizebox{0.3\linewidth}{!}{%
% \begin{tabular}{ll}
% \hline
% \hline
% Chart Type           & \textbf{mAP} \\
% \hline
% Horizontal Bar Chart & \textbf{0.953}     \\
% Vertical Bar Chart   & \textbf{0.972}     \\
% Line Plot            & \textbf{0.960}     \\
% Scatter Plot         & \textbf{0.925}     \\
% \hline
% \end{tabular}}
% \end{table}

\section{Conclusion} \label{sec:conc}
This work marks a significant contribution to the challenging task of chart information extraction by proposing a deep learning framework for each vital step in the process. The experimentation results validate that our proposed framework %outperforms SOTA (old) 
performs very good, well alongwith other SOTA in each step with F1-scores of 0.97 and 0.91 for chart-type and text-role classification, and 0.95 mAP for text detection, respectively. We also propose a solution for low resolution small-text recognition challenge by introducing an image enhancement step using ESRGANs. The most interesting aspect of this study is the use of positional information and relational dependency between multiple objects as a supporting factor for decision making in chart-type and text-role classification by utilizing hierarchical vision transformers. Due to this, we are able to propose a generic framework for multiple chart types i.e. horizontal bar, vertical bar, line plot and scatter plot. 

\bibliographystyle{IEEEtran}
\bibliography{references}

\end{document}